\documentclass{article}

\usepackage[sglblindworkshop,final]{neurips_2025}
\workshoptitle{MLxOR: Mathematical Foundations and Operational Integration of Machine Learning for Uncertainty-Aware Decision-Making}

\usepackage[utf8]{inputenc} 
\usepackage[T1]{fontenc}    
\usepackage{hyperref}       
\usepackage{url}            
\usepackage{booktabs}       
\usepackage{amsfonts,amsmath,amssymb}       
\usepackage{nicefrac}       
\usepackage{microtype}      
\usepackage{xcolor,graphicx}     
\usepackage{algorithm,algpseudocode}
\usepackage{bm} 
\newcommand{\bw}{\bm{w}}   
\usepackage{comment}
\title{SOCRATES: Simulation Optimization with Correlated Replicas and Adaptive Trajectory Evaluations}

\author{%
\parbox{\textwidth}{\centering\bfseries
\begin{tabular}{cccc}
Haoting Zhang$^{1,3}$ & Haoxian Chen$^{2,3}$ & Donglin Zhan$^{2}$ & Hanyang Zhao$^{2}$
\end{tabular}\\[0.32ex]
\begin{tabular}{cccc}
Henry Lam$^{2}$ & Wenpin Tang$^{2}$ & David Yao$^{2}$ & Zeyu Zheng$^{1}$
\end{tabular}}
\\[3.2ex]
$^{1}$University of California, Berkeley \qquad $^{2}$Columbia University \qquad$^{3}$Amazon \\ \\
\texttt{\{haoting\_zhang, zyzheng\}@berkeley.edu} \\
\texttt{\{hc3136, donglin.zhan, hz2684, henry.lam, wt2319, yao\}@columbia.edu}
}

\begin{document}

\maketitle

\begin{abstract}

The field of simulation optimization (SO) encompasses various methods developed to optimize complex, expensive-to-sample stochastic systems. Established methods include, but are not limited to, ranking-and-selection for finite alternatives and surrogate-based methods for continuous domains, with broad applications in engineering and operations management. The recent advent of large language models (LLMs) offers a new paradigm for exploiting system structure and automating the strategic selection and composition of these established SO methods into a tailored optimization procedure. This work introduces SOCRATES (Simulation Optimization with Correlated Replicas and Adaptive Trajectory Evaluations), a novel two-stage procedure that leverages LLMs to automate the design of tailored SO algorithms. The first stage constructs an ensemble of digital replicas of the real system. An LLM is employed to implement causal discovery from a textual description of the system, generating a structural `skeleton' that guides the sample-efficient learning of the replicas. In the second stage, this replica ensemble is used as an inexpensive testbed to evaluate a set of baseline SO algorithms. An LLM then acts as a meta-optimizer, analyzing the performance trajectories of these algorithms to iteratively revise and compose a final, hybrid optimization schedule. This schedule is designed to be adaptive, with the ability to be updated during the final execution on the real system when the optimization performance deviates from expectations. By integrating LLM-driven reasoning with LLM-assisted trajectory-aware meta-optimization, SOCRATES creates an effective and sample-efficient solution for complex SO optimization problems.
\end{abstract}

\section{Introduction}
\label{sec:intro}
\begin{figure}
    \centering
    \includegraphics[width=0.95\linewidth]{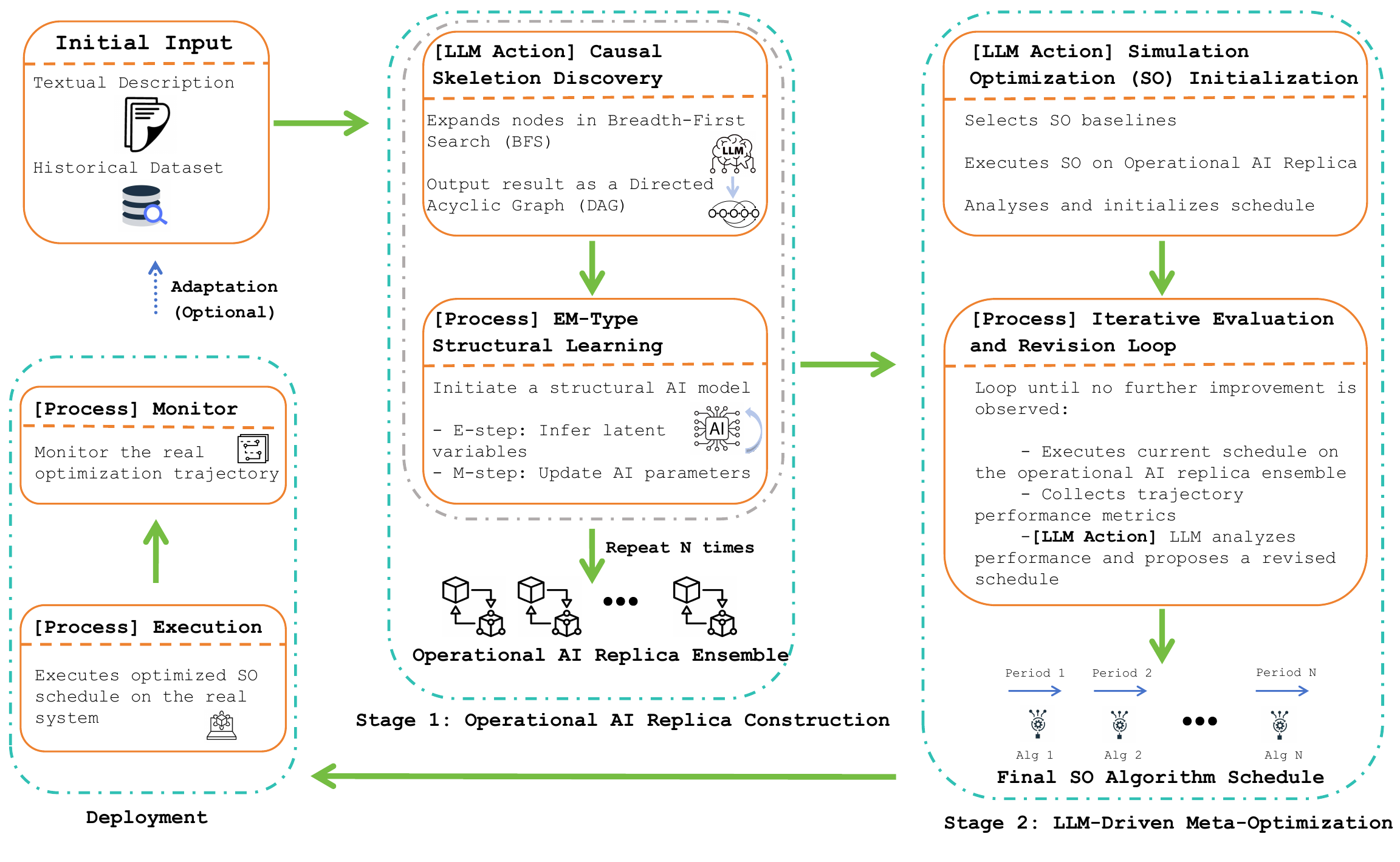}
    \caption{Pipeline of SOCRATES: Stage 1 learns an ensemble of operational AI replicas; Stage 2 employs these inexpensive replicas to evaluate and revise SO algorithms. The final SO algorithm is tailored to the real system before deployment, and an online adaptation is feasible when needed.}
    \vspace{-22pt}
    \label{fig:procedure}
\end{figure}

Stochastic systems can exhibit complex structures where system performance is not an analytical function of the decision variables but rather a surface that can only be evaluated through noisy samples. In many applications, the goal is to search the feasible set of decision variables in order to optimize system performance, which motivates the field of simulation optimization (SO) by regarding collecting system samples as simulations \cite{eckman2023simopt,peng2025review}. For problems with a finite set of decisions, ranking-and-selection procedures are designed to efficiently allocate simulation budgets to identify the best decision with statistical confidence \cite{ni2017efficient,hong2021review,keslin2024ranking,bai2025calibrating}. When the decision variables are within in a continuous space, surrogate model-based algorithms are employed to approximate the system performance to optimize \cite{barton2009simulation,zhang2021neural,hong2021surrogate}; a leading representative is Bayesian optimization \cite{frazier2018bayesian,cakmak2020bayesian,chen2023pseudo,zhan2025collaborative}, which employs a probabilistic model, such as a Gaussian process \cite{chen2013enhancing,barton2014quantifying,ryzhov2016convergence,zhang2023bayesian,zhang2023contextual,li2023inference}, to guide the optimization in a sample-efficient manner. Additionally, for highly non-structural and complex system performance, meta-heuristics methods provide a flexible and scalable framework \cite{zhou2013sequential,hu2014model}. On the other hand, most SO methods remain general-purpose black-box procedures. Because samples are costly and noisy, there has been comparatively limited work on tailoring SO algorithms to specific problem instances or domains, leaving per-instance adaptation underexplored.

These open challenges create an opportunity for large language models (LLMs) to inject structure and prior knowledge into the SO pipeline. A key frontier involves leveraging LLMs' extensive pre-training and semantic reasoning abilities to translate natural language problem descriptions into optimization components. Research in this area has progressed from foundational benchmarks that parse problem statements into mathematical representations \cite{nl4opt2022} to advanced multi-agent frameworks that break down the modeling pipeline into specialized roles like formulation, planning, and code generation \cite{thind2025optimai}. Concurrently, a significant research trend focuses on fine-tuning LLMs on domain-specific data to create specialized optimization assistants, enhancing both performance and data privacy \cite{orlm2024,jiang2025llmopt}. Beyond formulation and algorithm execution supported by LLMs, another paradigm employs LLMs as engines for methodological discovery, where program-synthesis and evolutionary loops produce verifiable new algorithms and heuristics \cite{wu2024survey}. Such approaches have yielded novel discoveries in mathematical sciences and improved heuristics for complex optimization problems by evolving semantic reasoning with executable implementations under automated evaluation \cite{funsearch2024,eoh2024icml,surina2025evotune}.  

Building on these opportunities, we propose \emph{SOCRATES (Simulation Optimization with Correlated Replicas and Adaptive Trajectory Evaluations)}—an LLM-driven meta-optimizer \cite{krus2003optimizing} that evaluates and refines SO algorithms prior to deployment on complex, expensive-to-sample stochastic systems. Our procedure has two stages. First, we use an LLM’s reasoning ability to infer a structural ``skeleton’’ from the system’s textual description, which guides data-efficient, structure-aware learning of an ensemble of system replicas (Sec. \ref{sec.twin}). This learning supports flexible selection of AI models and methods; we term the replicas \textit{Operational AI Replicas (OARs)}. These OARs have an affinity to
digital twins \cite{wang2020simulation,matta2023digital}, yet are purpose-built for evaluating and revising operations prior to deployment. In the second stage (Sec. \ref{sec.meta}), using the learned replicas as a testbed, we iteratively invoke LLMs as a meta-optimizer to analyze optimization-trajectory ensembles and select and refine a final SO algorithm. The resulting SO algorithm is tailored to the problem’s structure before deployment on the expensive real system. Finally, the procedure is adaptive: during execution on the real system, if the observed trajectory deviates from expectation, the twin ensemble can be updated and the optimization strategy revised for the remaining budget. A similar adaptive-feedback loop appears in Rich Preference Optimization (RPO) \cite{zhao2025fine}, where the system iteratively critiques, edits, and relabels preferences to refine a diffusion model. The entire pipeline is summarized in Figure \ref{fig:procedure}.

\section{Methodology}
\label{sec:method}


\subsection{OAR Construction via LLM–Guided Causal Skeleton and EM Learning}
\label{sec.twin}

\textbf{Problem setting.}
We build an OAR from two inputs:
(i) a textual specification $\mathcal{P}$ with system components, mechanism, constraints, etc. in natural language and
(ii) a historical dataset $\mathcal{H}=\{(x_i,y_i)\}_{i=1}^N$ of controllable inputs $x$ and observed objective $y$.
A key challenge is the mismatch between \emph{system complexity} and \emph{data scarcity} (limited observations for expensive sampling).
To address this, we 1) leverage LLMs to build a causal `skeleton' that represents the system structure and 2) implement a expectation-maximization (EM)-type data-efficient learning procedure:
\subsubsection{Step 1: Causal Skeleton Inference with LLMs}
We first use LLMs to infer a \emph{causal skeleton} $G=(V,E)$ over the system variables listed in the textual description $\mathcal{P}$. Here, $V$ is the set of variables listed in $\mathcal{P}$, $E$ denotes the set of edges that represent the causal relations between variables $v\in V$, and $G$ is a directed acyclic graph (DAG) that summarizes the causal relations and mechanism of the system. To infer the DAG from the textual description $\mathcal{P}$, a  breadth-first search (BFS) causal discovery procedure over variables $V$ involves the following steps: \textbf{1. Initialization:} Prompt the LLM with all variable names/descriptions and ask it to select the set of \emph{exogenous} variables (not caused by any others). These seed a BFS queue. \textbf{2. Expansion:} select a node $u$ from the queue and ask the LLM to select all variables that are \emph{caused by} $u$. \textbf{3. Insertion:} Add edges $(u\!\to\!v)$ that do not create cycles; push newly discovered children $v$ to the queue. Repeat until the queue is empty; see \cite{jiralerspong2024efficient}.

To be more specific, for the variables, we let $V = X \cup Z \cup \{Y\}$ denote inputs, latent internal components, and the objective. In our implementation, the result of the initialization step is fixed to be $X$ since they are controllable. Additionally, for the objective $Y$, we do not expand nodes since it is the final output. This BFS formulation reduces the number of LLM queries from $O(|V|^2)$ pairwise judgments to $O(|V|)$ expansions, while a per-edge cycle check guarantees a DAG. In addition, we include in the prompt a history of the inference, which accumulates previously discovered edges to improve consistency across turns of calling LLMs.


\subsubsection{Step 2: EM-Type Learning of Structural Mechanisms}
Given the skeleton $G$, we learn node-level mechanisms with a differentiable structural model.
Recall that $V = X \cup Z \cup \{Y\}$ denote inputs, latent internal components, and the objective.
For each non-input node $j\in Z \cup \{Y\} $ with parents $\mathrm{pa}(j)$ in $G$, we let
$$
\mathbf{s}_j \;=\; f_j\!\left( \mathbf{s}_{\mathrm{pa}(j)};\,\theta_j \right) \,+\, \boldsymbol{\varepsilon}_j,
\quad\text{with}\quad
Y \;=\; f_Y\!\left( \mathbf{s}_{\mathrm{pa}(Y)};\,\theta_Y \right) \,+\, \varepsilon_Y,
$$
where $\mathbf{s}_j \in \mathbb{R}^{d_j}$ is the $d_j$-dimensional vector state of node $j$, and $\mathbf{s}_{\mathrm{pa}(j)}$ is the concatenated vector state of its parents. The objective $Y$ remains a scalar. The map $f_j$ is approximated by AI models (e.g., an MLP) with parameters $\theta_j$ that outputs a $d_j$-dimensional vector.
We learn parameters $\theta=\{ \theta_j \}$ and infer latent variables $Z=\{\mathbf{s}_z\}_{z\in Z}$ via an EM-type learning procedure that explicitly respects the causal skeleton $G$ inferred by LLMs and accounts for the uncertainty within system components (latent variables). The learning is composed of alternating E/M steps as follows:

\textbf{E-step (Latent Inference).}
With the parameters of the map $\theta$ fixed, we infer latent variables $\widehat{Z}$ by minimizing a loss that trades off end-to-end objective fit and consistency with the current mechanisms:
\[
\mathcal{L}_E(\widehat{Z};\theta)
\;\;=\;\;
\frac{1}{N}\sum_{i=1}^N
\underbrace{\bigl\|y_i - \widehat{y}(x_i,\widehat{Z}_i;\theta)\bigr\|_2^2}_{\text{objective fit}}
\;+\;
\lambda\cdot \sum_{j\in Z} \frac{1}{N}\sum_{i=1}^N
\underbrace{\bigl\| \widehat{\mathbf{s}}_{j,i} - f_j(\mathbf{s}_{\mathrm{pa}(j),i};\theta_j) \bigr\|_2^2}_{\text{mechanism consistency}}.
\]
Here $\widehat{y}$ is the forward pass through $G$, using input $x_i$ and inferred latent variables $\widehat{Z}_i = \left\{\widehat{\mathbf{s}}_{j,i}, j\in Z\right\}$.
The term mechanism consistency keeps the E-step from drifting far from the current mechanisms and empirically stabilizes training in low-data regimes.

\textbf{M-step (Mechanism Update).}
With $\widehat{Z}$ fixed, we refit each component map $f_j(\,\cdot\,;\theta_j)$ by minimizing:
\[
\mathcal{L}_M(\theta;\widehat{Z})
\;\;=\;\;
\sum_{j\in Z} \frac{1}{N}\sum_{i=1}^N
\bigl\| \widehat{\mathbf{s}}_{j,i} - f_j(\mathbf{s}_{\mathrm{pa}(j),i};\theta_j) \bigr\|_2^2
\;+\;
\gamma \cdot \frac{1}{N}\sum_{i=1}^N
\bigl\|y_i - \widehat{y}(x_i,\widehat{Z}_i;\theta)\bigr\|_2^2.
\]

Lastly, to account for the uncertainty in LLM inference and EM-type learning procedure, we repeat the procedure multiple times independently to get a set of learned OARs. We then select top-$K$ of them based on the accuracy, and then apply an ensemble method to decide a final OAR with an optimal ensemble weight, enhancing the resilience of the subsequent operations.




\subsection{LLM-based Trajectory–Aware Meta Optimizer}
\label{sec.meta}

\textbf{Goal.}
Given a fixed OAR (Sec.~\ref{sec.twin}) and a fixed evaluation budget $T$, our objective is to select and revise the SO algorithm that will be applied to optimizing the real system, i.e., we implement a meta optimization (optimizing the optimizer) procedure on an OAR instead of directly applying an optimization algorithm to the real system. Compared to classical meta–optimizers that rely solely on quantitative evaluations, our method leverages LLMs to reason jointly over \emph{quantitative performances} (trajectory-based metrics) and \emph{qualitative insights}, integrating the LLM's pre-trained knowledge of algorithmic strategies with its ability to reason semantically about the optimization process.

Let $\mathcal{A}=\{a_1,\dots,a_m\}$ denote a set of baseline SO algorithms (e.g., variants of Bayesian optimization algorithms and meta–heuristics). Our procedure finally provides a schedule $\pi=\left((a_{j},\,T_{j})\right)_{j=1}^J$, where $a_j\in\mathcal{A}$ is executed for $T_j$ budgets. That is, the final SO algorithm is composed of different SO baseline algorithms implemented in different periods, leveraging the strengths of each algorithm at different stages of the optimization. Specifically, the procedure is composed of:

\textbf{1. Initialization:} For the real system with the textual description $\mathcal{P}$ and the budget $T$, we use an LLM to select several baseline SO algorithms based on its knowledge base and a predefined selection logic. Each selected SO algorithm is then implemented on the OAR for the full budget to collect its entire optimization trajectory. Finally, the LLM decides an initial schedule $\pi^{(0)}$ by analyzing the collected trajectories, utilizing its semantic understanding and reasoning ability.

\textbf{2. Iterative Revision:} Starting from an initial schedule, we employ LLMs to automatically and iteratively revise it. In each iteration, we: 1) implement the current schedule on the OAR, 2) collect trajectories and the associated performance metrics, and 3) input the current schedule and its performance into an LLM to request a revision. The LLM also has access to the performance of baseline SO algorithms for reference. To facilitate directional and efficient revisions, we include the revision history in the LLM’s input. If a newly proposed schedule does not outperform the previous one, we retain the earlier version but still append the new (inferior) schedule to the revision history. This allows the LLM to learn from both ``successful'' and ``unsuccessful'' attempts. The iterative revision process terminates when no further improvement is observed across successive iterations. 


Additionally, we evaluate and select schedules using an \emph{ensemble} of OARs (top-$K$ models from Sec.~\ref{sec.twin}). In this way, the final SO algorithm accounts for uncertainty arising from the OAR construction, thereby achieving resilience against discrepancies between the real system and the learned OAR. Furthermore, the ensemble enables us to construct a family of twins by assigning different ensemble weights. By regarding each OAR with a specific ensemble weight as a data point that provides a ``loss'' to minimize, the iterative revision procedure in our method can be cast as a learning epoch within the classical machine learning (ML) framework. In this manner, LLMs provide a ``semantic gradient'' to revise the input schedule of SO algorithms. That is, we integrate the iterative revision pipeline into a standardized ML framework that combines both the reasoning and semantic understanding of LLMs with the rigor of well-established ML training methodologies, ensuring both performance and generalizability.


In addition to the iterative revision on OARs, we implement an online \emph{adaptation} procedure during the final execution on the real system. We monitor the real optimization trajectory and the performance metrics. If this trajectory exhibits unexpected behavior compared to the performance simulated on OARs, we update the optimal ensemble weight with the new real-system data and call LLMs to iteratively revise the schedule for the remaining budget. This allows the meta-optimizer to adapt to potential mismatches between the learned OAR and the real system, further improving the data-efficiency of the final optimization. The detailed procedure of SOCRATES, as well as the experimental results, is included in the appendix.






\newpage
\bibliographystyle{abbrv}
\bibliography{references.bib}

\newpage
\appendix

\section{Detailed Methodology}
Here we include the detailed procedure of SOCRATES as a complement of Sec. \ref{sec:method}.

\subsection{Operational AI Replica Construction}
During the first stage of SOCRATES, we construct an ensemble of operational AI replicas (OARs), which leverage LLMs to 1) approximate the system under different ``what-if'' scenarios, and 2) provide a cost-effective ``testbed'' to evaluate and revise simulation optimization (SO) algorithms before execution on the expensive real system. The construction of OARs takes two types of input:
\begin{enumerate}
    \item A textual specification $\mathcal{P}$ with system components, mechanism, constraints, etc., in natural language;
\item A historical input/output (I/O) dataset $\mathcal{H}=\{(x_i,y_i)\}_{i=1}^N$ of controllable inputs $x$ and observed objective $y$.
\end{enumerate}

With these two sources of input, the construction of OARs begins with the causal relation discovery of the system components to \emph{qualify} the structure of the system. The inferred causal relations serve as the ``skeleton'' of OAR, and we then implement an expectation-maximization (EM) procedure to \emph{quantify} the dependencies between components in OAR.

\subsubsection{Causal Relation Inference}

Let $G=(V,E)$ denote the directed acyclic graph (DAG) to be inferred from a textual description $\mathcal{P}$ of the variables. We partition the variables as $V = X \cup Z \cup \{Y\}$, where $X$ are system inputs (the decision variables for SO), $Z$ are latent internal components, and $Y$ is the system input (the objective value for SO). The implementation employs the breadth–first search (BFS) procedure: 1) Initialization, 2) Expansion, and 3) Insertion with cycle checks:
\begin{enumerate}
    \item \textbf{Initialization:} The procedure begins with deciding the set of variables $V = X \cup Z \cup \{Y\}$:
    \begin{itemize}
\item A set $V$ of variable names, each with a short natural-language description, extracted by an LLM from the textual description $\mathcal{P}$.
\item The designated objective $Y\in V$ (treated as a sink: no outgoing edges).
\item A subset $X\subseteq V$ of exogenous variables. When provided, edges \textbf{into} $X$ are disallowed so that $X$ remains exogenous, i.e., $x\in X$ is not dependent on any other $v'\in V$.
\end{itemize}
After deciding the variable set $V$, we initialize the state of DAG:
\begin{itemize}
\item A directed graph $G$ initialized with node set $V$ and empty edge set $E$.
\item A list that tracks the variables (nodes of the graph) that have been visited to ensure each node is expanded at most once. We also record a \emph{queue} that decides the order of nodes to be expanded, which is initialized with the set of exogenous variables $X$.
\item A causal discovery history recording the instructions summarized by an LLM from the textual description $\mathcal{P}$, the variable catalog $\{ \langle v\rangle: \text{description}(v) \}_{v\in V}$, and, after each turn, the decided causal relation (the inserted edge of the graph) $(u\!\to\!v)$; this history is fed back into subsequent prompts input to LLM to improve consistency across turns. 
\end{itemize}

\item \textbf{Expansion (LLM query at a node):} We decide the expansion node by node. While the queue is non-empty, select a node $u$ from it (first-in-first-out). The prompt into an LLM to ask for expansion contains:
\begin{enumerate}
\item The list of exogenous variables.
\item A causal discovery history recording the general instruction, the variable description, and the currently discovered causal relationships to date. The history is dynamically compacted by an LLM when the length exceeds the pre-specified threshold.
\item The instruction for the current node $u$ for expansion: \emph{``Select ALL variables that are caused by $\langle u\rangle$.''}
\end{enumerate}
The output of LLM is restricted (through the prompt) to return a set of variable names from $V$, which is then used as the proposed children of $u$. 

\item \textbf{Insertion with checks:} For each proposed child $v$ of $u$, the implementation executes the following admissibility checks before inserting the edge $(u\!\to\!v)$:
\begin{enumerate}
\item {Sink constraint:} if $u=Y$, reject (enforces that $Y$ has no outgoing edges).
\item {Exogeneity constraint:} if $v\in X$, reject (keeps $X$ exogenous).
\item {Acyclicity:} tentatively add $(u\!\to\!v)$; if this creates a directed cycle, revert and reject.
\end{enumerate}
If $(u\!\to\!v)$ is accepted, then $v$ is appended to the queue if it has not yet been visited or enqueued. The causal discovery history is updated as well. This per-edge cycle check guarantees that the final inferred causal `skeleton' $G$ is DAG. In most scenarios, the proposed children of $u$ pass the insertion since all the requirements (e.g, acyclicity) and the information ($X$, $Y$, and previously-visited nodes, etc.) are provided in the prompt to LLM. On the other hand, LLMs cannot always strictly follow all the requirements, especially when the causal discovery history is compacted, and therefore these admissibility checks are necessary.

\end{enumerate}
The BFS procedure terminates when the queue is empty. The complexity in LLM calls is $O(|V|)$ expansions since each node is expanded at most once. Additionally, we consider the scenarios when only the I/O historical data is available. Since I/O is pre-fixed during the causal relation discovery in our implementation, the I/O data is not utilized, i.e., the causal relation discovery is based on the domain knowledge provided in the textual description $\mathcal{P}$ and the knowledge base of LLMs attained in extensive pre-training. In some scenarios when the data of intermediate variables $Z$ is available, the Expansion step augments the prompt to LLMs by providing the statistics in data (e.g. Pearson correlations) between the current node $u$ and other variables. This strategy has the potential to provide additional instructions that can help the LLM when domain text alone is ambiguous.

\subsubsection{EM-type Learning}
Given the LLM‑inferred causal skeleton $G=(V,E)$, we learn an OAR by attaching a learnable mechanism $f_j(\cdot;\theta_j)$ to each non‑input node $j\in Z\cup{Y}$, with inputs given by the states of its parents $\mathrm{pa}(j)$ in $G$. Written in vector form,
\begin{equation*}
    \mathbf{s}_j=f_j\left(\mathbf{s}_{\mathrm{pa}(j)} ; \theta_j\right)+\boldsymbol{\varepsilon}_j, \quad Y=f_Y\left(\mathbf{s}_{\mathrm{pa}(Y)} ; \theta_Y\right)+\varepsilon_Y .
\end{equation*}
The collection $f(\cdot;\theta)={f_j(\cdot;\theta_j)}_{j\in Z\cup{Y}}$ defines the OAR. The \emph{model family for each $f_j$ is flexible}: small MLPs, gradient‑boosted trees, monotone nets, or other light parametric learners can be selected per node based on variable type, smoothness, and known monotonicities. When a variable arrives in a raw, non‑numeric modality (categorical/text/ordinal, etc.), we introduce learned or library embeddings $e_j$ to map raw inputs into numeric states used by $f_j$. In practice, this \emph{per‑node model selection} can be automated by an AutoML layer \cite{he2021automl}, optionally seeded by LLM guidance; recent systems such as MLZero \cite{fang2025mlzero} demonstrate promising LLM‑augmented AutoML that we can leverage to pick reasonable $f_j$ families and embeddings without changing the EM loop.

We learn parameters $\theta$ and latent node variables $Z$ by alternating:
(i) an \emph{E‑step} that infers per‑sample latent states $\widehat{Z}$ with the mechanisms held fixed, balancing end‑to‑end objective fit and per‑node consistency; and
(ii) an \emph{M‑step} that refits the mechanisms to the inferred latents, with a small end‑to‑end term to keep global performance aligned.
Before the first E/M iteration, we perform a Stage‑0 end‑to‑end fit over the historical set that ignores explicit latents (i.e., forward only through $G$), which supplies an informed starting point $\theta^{(0)}$ and an initial set of latent states via a forward pass.

After we have set the model parameter $\theta$, the learning procedure of OAR is iterated by E, M steps, where we optimize latent variables $\widehat{Z}$ and the model parameters $\theta$ to minimize the loss functions respectively:
\begin{equation*}
    \begin{aligned}
&\mathcal{L}_E(\widehat{Z} ; \theta)=\frac{1}{N} \sum^N\left\|y_i-\widehat{y}\left(x_i, \widehat{Z}_i ; \theta\right)\right\|_2^2+\lambda \sum_{j \in Z} \frac{1}{N} \sum_{i=1}^N\left\|\widehat{\mathbf{s}}_{j, i}-f_j\left(\mathbf{s}_{\mathrm{pa}(j), i} ; \theta_j\right)\right\|_2^2,\\
&\mathcal{L}_M(\theta ; \widehat{Z})=\sum_{j \in Z} \frac{1}{N} \sum_{i=1}^{N}\left\|\widehat{\mathbf{s}}_{j, i}-f_j\left(\mathbf{s}_{\mathrm{pa}(j), i} ; \theta_j\right)\right\|_2^2+\gamma \frac{1}{N} \sum_{i=1}^N\left\|y_i-\widehat{y}\left(x_i, \widehat{Z}_i ; \theta\right)\right\|_2^2 .
\end{aligned}
\end{equation*}
Here $\lambda,\gamma>0$ trade local mechanism fidelity and global predictive performance. Additionally, since EM procedure is sensitive to the ``starting point,'' we therefore implement a multi-start learning procedure and select the best model based on a leave-out test set. The detailed procedure of learning OAR is included in Alg. \ref{alg:oar-em}.
\begin{algorithm}[t]
\caption{Structure-Aware OAR Learning: Stage-0 + EM with Multi-Start and Validation}
\label{alg:oar-em}
\begin{algorithmic}[1]
\Require Historical data $\mathcal{H} = \{(x_i, y_i)\}_{i=1}^N$; inferred causal skeleton $G = (V, E)$; train/val/test split ratios; EM rounds $R$; number of starts $S$.
\Ensure A learned OAR $f(\cdot\,; \theta^{\star})$.

\State \textbf{Split} $\mathcal{H}$ into $\mathcal{H}_{\text{train}}, \mathcal{H}_{\text{val}}, \mathcal{H}_{\text{test}}$ (fixed once per input).

\For{$s = 1$ \textbf{ to } $S$} \Comment{Multi-start restarts to reduce local minima}
    \State \textbf{Per-node model/embedding selection:} 
    \Statex \hspace{1.25em} For each $j \in Z \cup \{Y\}$, choose a candidate family $f_j \in \mathcal{F}_j$ and
    \Statex \hspace{1.25em} (if needed) encoder $e_j$ via rules or AutoML.
    
    \State \textbf{Initialize} $\theta^{(0,s)}$ (random or AutoML warm start); fix topology by $G$.
    
    \State \textbf{Stage-0 end-to-end fit:}
    \Statex \hspace{1.25em} Train $\theta$ on $\mathcal{H}_{\text{train}}$ to minimize:
    $$\frac{1}{|\mathcal{H}_{\text{train}}|} \sum_{(x,y) \in \mathcal{H}_{\text{train}}} |y - \hat{y}(x; \theta)|^2$$
    \Statex \hspace{1.25em} with early stopping on $\mathcal{H}_{\text{val}}$; set $\theta^{(0,s)} \gets$ best checkpoint.
    
    \State \textbf{Initialize latents} $\hat{Z}^{(0,s)}$: 
    \Statex \hspace{1.25em} Forward pass with $\theta^{(0,s)}$ on $\mathcal{H}_{\text{train}}$;
    \Statex \hspace{1.25em} For observed internals, clamp $\hat{\mathbf{z}}_j \gets e_j(\text{raw})$.
    
    \For{$t = 1$ \textbf{ to } $R$} \Comment{EM rounds}
        \State \textbf{E-step:} With $\theta^{(t-1,s)}$ fixed, update $\hat{Z}$ on $\mathcal{H}_{\text{train}}$ by
        \Statex \hspace{2.68em} $K_E$ gradient steps to reduce $\mathcal{L}_E(\hat{Z}; \theta^{(t-1,s)})$.
        \Statex \hspace{2.68em} Clamp known internals; project to feasible bounds if needed;
        \Statex \hspace{2.68em} warm-start from $\hat{Z}^{(t-1,s)}$.
        
        \State \textbf{M-step:} With $\hat{Z}^{(t,s)}$ fixed, refit $\theta$ on $\mathcal{H}_{\text{train}}$ to reduce
        \Statex \hspace{2.68em} $\mathcal{L}_M(\theta; \hat{Z}^{(t,s)})$ with early stopping on $\mathcal{H}_{\text{val}}$;
        \Statex \hspace{2.68em} set $\theta^{(t,s)} \gets$ best checkpoint this round.
        
        \State \textbf{Validation snapshot:} Compute
        $$\text{MSE}_{\text{val}}^{(t,s)} = \frac{1}{|\mathcal{H}_{\text{val}}|} \sum_{(x,y) \in \mathcal{H}_{\text{val}}} |y - \hat{y}(x, \hat{Z}^{(t,s)}; \theta^{(t,s)})|^2$$
        \Statex \hspace{2.68em} Keep the best $(\theta, \hat{Z})$ seen so far within this start.
        
        \If{no improvement for $\tau$ rounds} 
            \textbf{break} 
        \EndIf
    \EndFor
    
    \State \textbf{Per-start test score:} Evaluate the best checkpoint for start $s$ on $\mathcal{H}_{\text{test}}$;
    \Statex \hspace{1.25em} record $\text{MSE}_{\text{test}}^{(s)}$.
\EndFor

\State \textbf{Select} $s^{\star} = \arg\min_s \text{MSE}_{\text{test}}^{(s)}$ and return the corresponding OAR $f(\cdot\,; \theta^{\star})$.
\end{algorithmic}
\end{algorithm}

\paragraph{Why structure‑aware EM vs.\ pure end‑to‑end.}
Compared with a single unstructured end‑to‑end model (e.g, an MLP from $X$ to $Y$), the proposed learning is:
\begin{enumerate}
\item \textbf{Data‑efficient.} The LLM‑guided skeleton prunes the hypothesis space; the EM splits learning into local maps constrained by $G$, which substantially lowers sample complexity in scarce‑data regimes.
\item \textbf{Interpretable.} Each $f_j$ is a named, low‑dimensional mechanism attached to a node; latent variables $\widehat{Z}$ are explicit and can be inspected, constrained, or partially instrumented. Instead, an end-to-end model is a black box, of which the performance is difficult to explain. 
\item \textbf{Flexible to partial observability.} Whenever some internal components are measured (fully or sporadically), the E‑step can clamp or supervise those states, and the M‑step can fit $f_j$ directly without re‑architecting the entire model.
\end{enumerate}

To account for the uncertainty in LLM-based causal relation discovery and EM-type learning procedure, we implement the OAR construction $J$ times independently (repeating Alg. \ref{alg:oar-em} from scratch), and get a set of learned OARs:
\begin{equation*}
    \left\{\left ( \operatorname{OAR}_{1},\operatorname{MSE}_{1}   \right ) , \left ( \operatorname{OAR}_{2},\operatorname{MSE}_{2}   \right ),\ldots, \left ( \operatorname{OAR}_{J},\operatorname{MSE}_{J}   \right )  \right\},
\end{equation*}
where $\operatorname{MSE}_{j}$ represents the fidelity of $\operatorname{OAR}_{j}$ as in Alg. \ref{alg:oar-em}. Then we select a top-$K$ of OARs with the highest fidelity, i.e., 
\begin{equation*}
    \mathcal{K}^{\star}=\left\{k_{(1)}, \ldots, k_{\left(K\right)}\right\} \subset\{1, \ldots, J\}, \quad \operatorname{MSE}_{k_{(1)}} \leq \cdots \leq \operatorname{MSE}_{k_{\left(K\right)}},
\end{equation*}
and construct an ensemble:
\begin{equation*}
    \operatorname{OAR}^{\star} = \sum_{k\in\mathcal{K}^*}w_k^{\star} \operatorname{OAR}_k.
\end{equation*}
Here the optimal ensemble weight is decided by
    $w_k^{\star}  = \frac{\frac{1}{\operatorname{MSE}_{k}+\epsilon } }{\sum_{k'\in\mathcal{K}^\star }\frac{1}{\operatorname{MSE}_{k'}+\epsilon }}$
with a fixed constant $\epsilon>0$ to numerically stabilize the weight. In this manner, the ensemble of OARs exhibits the resilience against the uncertainty involved in the construction procedure, especially hallucinations from LLMs. 

In addition to resilience, the ensemble of OARs brings two additional advantages: First, by re-weighting the ensemble weights around the optimal weight $w^{\star}$, we have access to a \emph{family} of OARs, which allows for generating sufficient ``data points'' for the subsequent meta-optimization procedure. Second, the ensemble of OARs shares similar spirits with Bayesian model averaging (BMA). In this manner, when the I/O dataset is updated with new observations, the optimal ensemble weight of OARs can be updated efficiently instead of re-constructing the OARs from scratch. 


\subsection{Trajectory-Aware Meta Optimizer}
At the second stage of SOCRATES, we leverage LLMs to evaluate and revise SO algorithms, based on the ensembles of OARs learned at the first stage. Both the evaluation and revision are supported by the semantic reasoning of LLMs. In this manner, the update of SO algorithms proposed by LLMs can be regarded as ``semantic gradient \cite{yang2023large},'' and we then cast the second stage into a rigorous learning framework to guarantee the generalizability of the final SO algorithm.

\subsubsection{Dataset Construction for Meta-Optimization via OAR Ensembles}
\label{sec:dataset-oar-ensembles}
We begin with constructing a ``dataset'' for meta-optimization. Recall that, $\{\operatorname{OAR}_k,\ \mathrm{MSE}_k\}_{k\in\mathcal{K}^\star}$ denote the top-$K$ OARs. Let $\bw^\star=\left ( w^{\star}_1,w^{\star}_2,\dots,w^{\star}_K \right ) ^{\top}\in\Delta^{K-1}$ be the optimal ensemble weight on the probability simplex
\[
w_k^{\star}\ \propto\ \frac{1}{\mathrm{MSE}_k+\epsilon},\qquad 
\sum_{k\in\mathcal{K}^\star} w_k^\star = 1,\quad w_k^\star \ge 0,
\]
with a small $\epsilon>0$ for numerical stability. An OAR ensemble parameterized by any $\bw\in\Delta^{K-1}$ is the linear mixture
\[
\widetilde{\operatorname{OAR}}(\,\cdot\,;\bw)\ :=\ \sum_{k\in\mathcal{K}^\star} w_k\,\operatorname{OAR}_k(\,\cdot\,),
\]
which we regard as one ``data point'' for meta‑optimization. To facilitate a rigorous learning framework, we construct a data with $M$ data points. Below we specify: (i) how to \emph{sample} many such ensembles $\bw^{(m)}$ around $\bw^\star$; (ii) how to assign each a scalar \emph{importance weight} $\tilde{\omega}_m$; and (iii) how to obtain a weighted, stratified \emph{train/validation/test} split.
\begin{enumerate}

    \item \textbf{Sampling ensemble weights near $\bw^\star$:} We draw $M$ weight $K$-dimensional vectors $\{\bw^{(m)}\}_{m=1}^M$ from a mixture of $L$ Dirichlet distributions centered at $\bw^\star$ with varying concentration (``radius'') to balance \emph{local} exploration near $\bw^\star$ and \emph{global} coverage of the simplex:
$$\begin{aligned}
&\textstyle \bw^{(m)} \sim q(\bw)\ :=\ \sum_{\ell=1}^{L} \pi_\ell\,\mathrm{Dir}\!\left(\boldsymbol{\alpha}^{(\ell)}\right),
\qquad \boldsymbol{\alpha}^{(\ell)}\ :=\ \tau_{\ell}\,\bigl(\ (1-\delta K)\,\bw^\star + \delta \,\mathbf{1}\ \bigr), 
\label{eq:mixture-dirichlet}\\
&\textstyle \tau_1<\tau_2<\cdots<\tau_{L},\quad \sum_{\ell} \pi_{\ell}=1,\quad \delta\in(0,1/K), \nonumber
\end{aligned}$$
where $\tau_{\ell}$ controls concentration around $\bw^\star$ (large $\tau_{\ell}$ yields tight samples near $\bw^\star$; small $\tau_{\ell}$ explores farther), and a tiny floor $\delta$ prevents zero Dirichlet parameters when some $w_k^\star=0$. A simple and effective choice is a geometric ladder $\tau_{\ell}=\tau_{\min}\,\rho^{\,\ell-1}$ with $\rho>1$ and uniform $\pi_{\ell}=1/L$.

\item\textbf{Defining importance weights:}
Each sampled ensemble $\bw^{(m)}$ receives an importance weight $\tilde{\omega}_m$ considering two aspects:
(i) \emph{closeness} to the best ensemble $\bw^\star$, and
(ii) \emph{fidelity} implied by the base‑OAR MSEs.
We quantify closeness by the KL divergence on the simplex and approximate ensemble fidelity using MSE:
\[
D_m\ :=\ D_{\mathrm{KL}}\!\bigl(\bw^{(m)}\ \Vert\ \bw^\star\bigr), 
\qquad 
\widehat{\mathrm{MSE}}\bigl(\bw^{(m)}\bigr)\ :=\ \sum_{k\in\mathcal{K}^\star} w^{(m)}_k\,\mathrm{MSE}_k.
\]
We set the importance weight:
\begin{equation*}
\zeta_m\ :=\ \exp\!\Bigl(-\alpha\, D_m\Bigr)\ \cdot\ \Bigl(\widehat{\mathrm{MSE}}(\bw^{(m)})+\epsilon\Bigr)^{-\beta},
\label{eq:score-zeta}
\end{equation*}
with hyperparameters $\alpha,\beta>0$, and then normalize the weight as:
\begin{equation*}
\tilde{\omega}_m\ :=\ \frac{\zeta_m}{\sum_{j=1}^{M} \zeta_j}.
\label{eq:omega-tilde}
\end{equation*}

The dataset is the weighted set of ensemble OARs
\[
\mathcal{D}\ :=\ \bigl\{\, \bigl(\widetilde{\operatorname{OAR}}^{(m)},\ \bw^{(m)},\ \tilde{\omega}_m\bigr)\ \bigr\}_{m=1}^{M},
\qquad
\widetilde{\operatorname{OAR}}^{(m)}(\,\cdot\,)\ :=\ \sum_{k\in\mathcal{K}^\star} w_k^{(m)}\,\operatorname{OAR}_k(\,\cdot\,).
\]
In subsequent meta‑optimization, SO algorithms are run on subsets of $\mathcal{D}$ to generate trajectories; the performance of algorithms is aggregated using the weights $\tilde{\omega}_m$, while the full trajectories are retained to drive LLM‑based revisions.

\item  \textbf{Weighted, stratified train/validation/test split.}
We partition $\mathcal{D}$ into $(\mathcal{D}_{\mathrm{train}},\mathcal{D}_{\mathrm{val}},\mathcal{D}_{\mathrm{test}})$ with proportions $(\rho_{\mathrm{tr}},\rho_{\mathrm{val}},\rho_{\mathrm{te}})$ (e.g., $0.70/0.15/0.15$), ensuring both (i) coverage across distances to $\bw^\star$ and (ii) respect for the importance weights.

\begin{enumerate}
\item \textbf{Distance stratification.} Compute distances $D_m$ and define $S$ strata by weighted quantiles of $\{D_m\}_{m=1}^M$ under weights $\tilde{\omega}_m$; i.e., pick cutpoints $0=q_0<q_1<\cdots<q_S=1$ and choose thresholds $d_s$ such that
\[
\sum_{m:\, D_m \le d_s} \tilde{\omega}_m\ =\ q_s\sum_{m=1}^M \tilde{\omega}_m.
\]
Let $\mathcal{I}_s=\{m:\ d_{s-1}< D_m \le d_s\}$ be stratum $s$.

\item \textbf{Per‑stratum targets.} For each stratum $s$, compute total weight $W_s=\sum_{m\in\mathcal{I}_s}\tilde{\omega}_m$ and set target weights
\[
T_{s}^{\mathrm{tr}}=\rho_{\mathrm{tr}}\,W_s,\qquad
T_{s}^{\mathrm{val}}=\rho_{\mathrm{val}}\,W_s,\qquad
T_{s}^{\mathrm{te}}=\rho_{\mathrm{te}}\,W_s.
\]

\item \textbf{Weighted selection (without replacement).} Within each $\mathcal{I}_s$, sample indices into the train set sequentially \emph{without replacement} with probabilities proportional to $\tilde{\omega}_m$ until the accumulated weight $\sum_{m\in \mathcal{D}_{\mathrm{train}}\cap \mathcal{I}_s}\tilde{\omega}_m$ first exceeds $T_s^{\mathrm{tr}}$ (greedy acceptance with stochastic tie‑breaking). Repeat on the remaining pool for validation until exceeding $T_s^{\mathrm{val}}$; assign the rest in $\mathcal{I}_s$ to test. Optionally, to reduce variance, enforce a minimal count per stratum for each split (e.g., at least $2$ items) by adjusting the last few assignments with the smallest absolute deviation from the targets.
\end{enumerate}

This procedure yields $(\mathcal{D}_{\mathrm{train}},\mathcal{D}_{\mathrm{val}},\mathcal{D}_{\mathrm{test}})$ that (i) preserve weighted coverage across a continuum of closeness to $\bw^\star$ (near, mid, far shells), (ii) concentrate effort according to $\tilde{\omega}_m$, and (iii) avoid leakage by sampling without replacement. The train split is then used to drive iterative, trajectory‑based algorithm revision; the validation split is held out to decide when a revision overfits the training ensemble pool; the test split is kept fixed for final selection among candidate revised algorithms.

\end{enumerate}
 The construction of the dataset is flexible with different strategies of sampling ensemble weights, defining importance weights, and dataset split. The above serves as an example for illustration.

 \subsubsection{Learning Framework of LLM-Driven SO Evaluation \& Revision}
\label{sec:learn-llm-schedule}

Given the OAR‑ensemble dataset $\mathcal{D}=\{(\widetilde{\mathrm{OAR}}^{(m)},\bw^{(m)},\tilde{\omega}_m)\}_{m=1}^{M}$ from Sec.~\ref{sec:dataset-oar-ensembles}, a fixed SO budget $B$, and a library $\mathcal{A}$ of baseline SO algorithms, the goal is to learn a \emph{schedule} (piecewise execution plan)
\[
\pi \;=\; \left((a_{j},\,T_{j})\right)_{j=1}^{J},\qquad a_j\in\mathcal{A},\ \ T_j\in\mathbb{N},\ \ \sum_{j=1}^{J}T_j = B,
\]
that will be deployed on the real system. The learning proceeds in \emph{epochs}. Within an epoch, an LLM proposes \emph{trajectory‑aware revisions} of $\pi$ using (i) the full SO trajectories generated on OAR ensembles in the training split and (ii) a textual revision history accumulated \emph{within the current epoch} (to learn from both successful and unsuccessful attempts). Revisions are accepted only if they improve a weighted scalar score computed from per‑trajectory metrics. After several intra‑epoch revisions, the candidate schedule is validated on the validation split to detect overfitting; epochs with a large train–validation gap are rejected. The test split is fixed for the final schedule selection across multiple outer runs (distinct train/validation partitions and initializations). We summarize the learning framework in Alg. \ref{alg:socrates-traj}, and more details are included as follows:

\textbf{Setup and notation.} Let $\mathcal{D}_{\mathrm{train}},\mathcal{D}_{\mathrm{val}},\mathcal{D}_{\mathrm{test}}$ be the weighted splits of $\mathcal{D}$. For minimization (the maximization case follows by sign change), executing a schedule $\pi$ on an ensemble $\widetilde{\mathrm{OAR}}^{(m)}$ yields a trajectory up to time $T$
\begin{equation}
\label{eq:trajectory}
\mathsf{Traj}^{(m)}(\pi)\;=\;\left\{\left(x_t^{(m)},\,y_t^{(m)}\right)\right\}_{t=1}^{T},
\end{equation}
where $x_t^{(m)}$ is the decision variable and $y_t^{(m)}$ is the objective value.
The LLM revision operator is an abstract map
\[
\widehat{\pi}\;=\;\mathsf{Rev}_{\mathrm{LLM}}\!\left(\,\pi,\ \{\mathsf{Traj}^{(m)}(\pi)\}_{m\in\mathcal{I}},\ \mathcal{H}\,\right),
\]
which takes the current schedule $\pi$, a set of training trajectories indexed by $\mathcal{I}\subseteq\{1,\dots,M\}$, and a textual history $\mathcal{H}$ maintained \emph{within the epoch}, and returns a candidate schedule $\widehat{\pi}$. Acceptance uses a scalar score $S(\pi\mid\mathcal{D}')$ on a split $\mathcal{D}'$ (see Alg.~\ref{alg:scoreandlog}), with revision rule
\[
\pi\ \leftarrow\ \begin{cases}
\widehat{\pi}, & \text{if } S(\widehat{\pi}\mid\mathcal{D}_{\mathrm{train}})\ \ge\ S(\pi\mid\mathcal{D}_{\mathrm{train}}),\\
\pi, & \text{otherwise},
\end{cases}
\quad\text{and}\quad \mathcal{H}\ \leftarrow\ \mathcal{H}\ \cup\ \text{``(schedule,\ metrics,\ score)''},
\]
i.e., the schedule updates when the performance is better, while the history that includes the proposed schedule, the trajectory-based metrics, and the associated scalar score, is appended for every revision. At the end of $T_{\mathrm{rev}}$ intra‑epoch revisions, we compute $S(\pi\mid\mathcal{D}_{\mathrm{val}})$ and accept the epoch if $\big|S(\pi\mid\mathcal{D}_{\mathrm{train}})-S(\pi\mid\mathcal{D}_{\mathrm{val}})\big|\le \varepsilon_{\mathrm{gap}}$; otherwise we revert to the previous epoch’s $\pi$. Revision histories are \emph{not} carried across epochs to prevent unbounded context growth and to encourage exploration of revision strategies, while the performances of the baseline algorithms are always included in the history for LLMs to refer to.

\textbf{Trajectory metrics and score (minimization).}
For each OAR ensemble $m$ and schedule $\pi$, let the full trajectory be as in (\ref{eq:trajectory}), and define the 
best–so–far sequence $b_t^{(m)}:=\min_{1\le s\le t}y_s^{(m)}$. 
To compare across ensembles without relying on baseline quantiles, we normalize per trajectory using 
a one‑time ``heavy'' reference optimum $\widehat{y}_{\star}^{(m)}$ (obtained once per ensemble with a large 
budget, e.g., $10\!\times$–$20\!\times$ the standard budget). 
To avoid degeneracy when the initial gap is tiny, we adopt a robust range:

\[
\rho^{(m)} \;:=\; q_{0.9}\!\bigl(\{y_t^{(m)}\}_{t=1}^{T}\bigr)-q_{0.1}\!\bigl(\{y_t^{(m)}\}_{t=1}^{T}\bigr),
\qquad
\Delta_{\star}^{(m)} \;:=\; \max\!\Bigl\{\bigl|b^{(m)}_1-\widehat{y}_{\star}^{(m)}\bigr|,\; \rho^{(m)}\Bigr\}+\varepsilon,
\]
where $q_\alpha$ denotes the empirical $\alpha$‑quantile on the trajectory and $\varepsilon>0$ is small.

We compute the following metrics (each in $[0,1]$ by construction). We use $[z]_+ := \max\{z,0\}$.

\begin{itemize}
\item \textbf{Final improvement.} Normalized net improvement (clipped):
\[
I_{\mathrm{final}}^{(m)} \;:=\; 
\min\!\left\{1,\; \frac{\bigl(b^{(m)}_1-b^{(m)}_T\bigr)_{+}}{\Delta_{\star}^{(m)}}\right\}.
\]

\item \textbf{Any‑time area under the improvement curve.} Rewards earlier gains by averaging the 
normalized best–so–far improvement over time:
\[
\mathrm{AUC}_{\mathrm{any}}^{(m)} \;:=\; 
\frac{1}{T-1}\sum_{t=2}^{T}\min\!\left\{1,\; \frac{\bigl(b^{(m)}_1-b^{(m)}_{t}\bigr)_{+}}{\Delta_{\star}^{(m)}}\right\}.
\]

\item \textbf{Monotonicity on the raw trajectory (tolerance $\xi \ge 0$).} Fraction of non‑worsening raw steps:
\[
\mu_{\mathrm{raw}}^{(m)} \;:=\; \frac{1}{T-1}\sum_{t=2}^{T}\mathbb{I}\!\left\{\,y_t^{(m)} \le y_{t-1}^{(m)}+\xi\,\right\}.
\]

\item \textbf{Stability via overshoot above the incumbent.} Penalizes oscillation/backtracking relative to the 
incumbent; larger is better:
\[
\sigma_{\mathrm{osc}}^{(m)} \;:=\; 
1 \;-\; \min\!\left\{1,\; \frac{1}{(T-1)\,\Delta_{\star}^{(m)}}\sum_{t=2}^{T}\bigl(y_t^{(m)}-b_{t-1}^{(m)}\bigr)_{+}\right\}.
\]

\end{itemize}

Stacking the $P$ metrics into $\mathbf{u}^{(m)}(\pi)\in[0,1]^P$ ($P=4$ with the order above), 
the per‑ensemble scalar score is a convex combination
\begin{equation}
\label{score}
s_m(\pi) \;=\; \sum_{p=1}^{P}\lambda_p\,u^{(m)}_{p}(\pi),
\qquad
\lambda_p\ge 0,\ \ \sum_{p=1}^{P}\lambda_p=1.
\end{equation}
The weights $\lambda_p$ are \emph{user‑specified} to emphasize final performance vs.\ stability, or \emph{estimated offline from pairwise preferences} 
(e.g., Bradley–Terry/logistic regression) collected by querying an LLM on trajectory plots and metric tables.

Finally, the schedule score on a split $\mathcal{D}'\subseteq\mathcal{D}$ (train/val/test) is the importance‑weighted mean
\[
S(\pi\mid\mathcal{D}')
\;=\;
\frac{\sum_{(\widetilde{\mathrm{OAR}}^{(m)},\bw^{(m)},\tilde{\omega}_m)\in\mathcal{D}'} \tilde{\omega}_m\, s_m(\pi)}
{\sum_{(\widetilde{\mathrm{OAR}}^{(m)},\bw^{(m)},\tilde{\omega}_m)\in\mathcal{D}'} \tilde{\omega}_m}.
\]

\textbf{Why trajectories and why a schedule of different SO algorithms.}
Different SO algorithm families exhibit complementary and phase‑dependent behavior. For instance, a meta-heuristic may first be deployed for a global exploration epoch to identify multiple promising regions in a given landscape. The decision to switch to a different algorithm is guided by the optimization trajectory, which reveals phase-dependent behaviors that would be obscured by a single final metric. When the algorithm's progress stalls or its search becomes narrow, the process transitions to another one (e.g., Bayesian optimization (BO) for sample-efficient refinement) \cite{volz2019benchmarking,santoni2024comparison,kuudela2024performance}. Even within the BO class, different choices of surrogate models or acquisition functions exhibit varying performances across different phases. These phase transitions are reflected in \emph{trajectory features} which are invisible to a single end‑of‑run metric. By exposing the full trajectories and their metrics to the LLM, the revision operator can (i) detect regime changes (e.g., stalling monotonicity, rising oscillation, feasibility collapse), (ii) switch to an algorithm class whose inductive bias fits the current regime (e.g., BO for fine‑grained exploitation after early exploration, or a restart‑capable heuristic when BO’s uncertainty collapses prematurely), and (iii) allocate remaining budget across phases accordingly. Thus, \emph{trajectory} and \emph{schedule} mutually justify each other: trajectories provide the signals that inform switching, while a schedule integrates complementary strengths across phases to outperform a standalone baseline algorithm.

\begin{algorithm}[t]
\caption{Learning Framework of an SO Schedule on OAR Ensembles}
\label{alg:socrates-traj}
\begin{algorithmic}[1]
\Require Problem description $\mathcal{P}$; OAR ensemble dataset $\mathcal{D}=\{(\widetilde{\mathrm{OAR}}_m,\tilde{w}_m)\}_{m=1}^{M}$ with importance weights $\tilde{w}_m>0$; fixed test split $\mathcal{D}_{\mathrm{test}}\subset\mathcal{D}$; number of outer runs $R$; epochs per run $E$; intra‑epoch revision steps $T_{\mathrm{rev}}$; generalization threshold $\varepsilon_{\mathrm{gap}}\!>\!0$; budget $B$; the function $\Call{ScoreAndLog}{\pi,\,\mathcal{D}}$ defined in Alg.~\ref{alg:scoreandlog}.
\Ensure Final SO schedule $\pi^{\star}$ selected by performance on $\mathcal{D}_{\mathrm{test}}$.
\Statex

\State Fix $\mathcal{D}_{\mathrm{test}}$ once. Let $\mathcal{D}_{\setminus \mathrm{test}} \gets \mathcal{D}\setminus \mathcal{D}_{\mathrm{test}}$.
\For{$r=1$ \textbf{to} $R$} \Comment{Outer runs}
  \State Partition $\mathcal{D}_{\setminus \mathrm{test}}$ into disjoint $\mathcal{D}^{(r)}_{\mathrm{train}}$ and $\mathcal{D}^{(r)}_{\mathrm{val}}$, accounting for importance weights.
  \Statex

  \State \textbf{(Baseline reference)} Query an LLM with $\mathcal{D}$ to select baseline algorithms $\mathcal{A}^{(r)}$.
  \State For each $a\in\mathcal{A}^{(r)}$ and each $(\widetilde{\mathrm{OAR}}_m,\tilde{w}_m)\in\mathcal{D}^{(r)}_{\mathrm{train}}$, run $a$ for budget $B$, compute trajectory metrics, and record a per‑ensemble, natural‑language reference $\mathcal{R}^{(0)}$.
  \Statex

  \State \textbf{(Initialization)} Obtain a schedule $\pi^{(0)}$ from an LLM using $\mathcal{D}$ and $\mathcal{R}^{(0)}$; set $\pi_{\mathrm{acc}}\!\gets\!\pi^{(0)}$.
  \For{$e=1$ \textbf{to} $E$} \Comment{Learning epochs}
    \State $\mathcal{H}^{(e)} \gets\mathcal{R}^{(0)}$;\quad $\pi^{(e,0)} \gets \pi_{\mathrm{acc}}$.
    \State $S^{(e,0)}_{\mathrm{train}},\ \{\mathbf{u}^{(m)}(\pi^{(e,0)})\},\ \mathsf{Traj}^{(e,0)}
           \gets \Call{ScoreAndLog}{\pi^{(e,0)},\,\mathcal{D}^{(r)}_{\mathrm{train}}}$.
    \State Append to $\mathcal{H}^{(e)}$ a concise record of $\pi^{(e,0)}$, per‑ensemble metrics, and $S^{(e,0)}_{\mathrm{train}}$.
    \For{$t=1$ \textbf{to} $T_{\mathrm{rev}}$} \Comment{Trajectory‑guided revisions}
      \State \textbf{LLM proposal:}
        $\widehat{\pi}^{(e,t)} \gets \mathsf{Rev}_{\mathrm{LLM}}\!\big(\,
          \pi^{(e,t-1)},\
          \mathsf{Traj}^{(e,t-1)},\
          \mathcal{H}^{(e)}
        \big)$.
      \State $S^{\mathrm{cand}}_{\mathrm{train}},\ \{\mathbf{u}^{(m)}(\widehat{\pi}^{(e,t)})\},\ \mathsf{Traj}^{\mathrm{cand}}
             \gets \Call{ScoreAndLog}{\widehat{\pi}^{(e,t)},\,\mathcal{D}^{(r)}_{\mathrm{train}}}$.
      \State Append to $\mathcal{H}^{(e)}$ a record of the candidate schedule, per‑ensemble metrics, and $S^{\mathrm{cand}}_{\mathrm{train}}$.
      \If{$S^{\mathrm{cand}}_{\mathrm{train}} > S^{(e,t-1)}_{\mathrm{train}}$} \Comment{Improvement}
        \State $\pi^{(e,t)} \gets \widehat{\pi}^{(e,t)}$;\quad $S^{(e,t)}_{\mathrm{train}} \gets S^{\mathrm{cand}}_{\mathrm{train}}$;\quad
               $\mathsf{Traj}^{(e,t)} \gets \mathsf{Traj}^{\mathrm{cand}}$
      \Else
        \State $\pi^{(e,t)} \gets \pi^{(e,t-1)}$;\quad $S^{(e,t)}_{\mathrm{train}} \gets S^{(e,t-1)}_{\mathrm{train}}$;\quad
               $\mathsf{Traj}^{(e,t)} \gets \mathsf{Traj}^{(e,t-1)}$
      \EndIf
    \EndFor
    \State \textbf{Validation:}\quad
           $S^{(e)}_{\mathrm{val}},\ \_,\ \_ \gets \Call{ScoreAndLog}{\pi^{(e,T_{\mathrm{rev}})},\,\mathcal{D}^{(r)}_{\mathrm{val}}}$;\quad
           $S^{(e)}_{\mathrm{train}} \gets S^{(e,T_{\mathrm{rev}})}_{\mathrm{train}}$.
    \If{$\big|S^{(e)}_{\mathrm{train}} - S^{(e)}_{\mathrm{val}}\big| \le \varepsilon_{\mathrm{gap}}$} \Comment{Accept epoch}
      \State $\pi_{\mathrm{acc}} \gets \pi^{(e,T_{\mathrm{rev}})}$
    \EndIf
  \EndFor
  \State $\pi^{(r)}_{\mathrm{final}}\!\gets\!\pi_{\mathrm{acc}}$;\quad
        $S^{(r)}_{\mathrm{test}},\ \_,\ \_ \gets \Call{ScoreAndLog}{\pi^{(r)}_{\mathrm{final}},\,\mathcal{D}_{\mathrm{test}}}$.
\EndFor
\State Select $r^{\star} \!\in\! \arg\max_{r} S^{(r)}_{\mathrm{test}}$ and \textbf{return} $\pi^{\star}\!=\!\pi^{(r^{\star})}_{\mathrm{final}}$.
\end{algorithmic}
\end{algorithm}

\begin{algorithm}[t]
\caption{ScoreAndLog: evaluate a schedule and return score, per‑ensemble metrics, and trajectories}
\label{alg:scoreandlog}
\begin{algorithmic}[1]
\Function{ScoreAndLog}{$\pi,\,\mathcal{D}'$} \Comment{$\mathcal{D}'\subseteq\mathcal{D}$}
  \ForAll{$(\widetilde{\mathrm{OAR}}_m,\,\tilde{w}_m)\in\mathcal{D}'$}
    \State Run $\pi$ on $\widetilde{\mathrm{OAR}}_m$ for budget $B$; log $\mathcal{T}^{(m)}(\pi)=\{(x^{(m)}_t,y^{(m)}_t)\}_{t=1}^{B}$.
    \State Compute trajectory metrics $\mathbf{u}^{(m)}(\pi)\in[0,1]^P$ (Sec.~\ref{sec:learn-llm-schedule}); set $s_m \gets \sum_{p=1}^{P}\lambda_p\,u^{(m)}_{p}(\pi)$.
  \EndFor
  \State $S \left(\pi\mid\mathcal{D}'\right)\gets \dfrac{\sum_{(\widetilde{\mathrm{OAR}}_m,\tilde{w}_m)\in\mathcal{D}'} \tilde{w}_m\, s_m}{\sum_{(\widetilde{\mathrm{OAR}}_m,\tilde{w}_m)\in\mathcal{D}'} \tilde{w}_m}$\Comment{Calculate the score}
  \State \textbf{return} $S$, $\{\mathbf{u}^{(m)}(\pi)\}_{m\in\mathcal{D}'}$, and $\{\mathcal{T}^{(m)}(\pi)\}_{m\in\mathcal{D}'}$
\EndFunction
\end{algorithmic}
\end{algorithm}

\section{Experiments}
\subsection{Multi-SKU Single-Echelon Warehouse and Base-Stock Optimization}
\label{subsec:warehouse-bs}
We consider a system in supply chain management. To be more specific, the system is a single-echelon warehouse that manages multiple products day by day. Each product has a fixed replenishment lead time, a holding cost for items kept on the shelf, and a backorder penalty for unfilled demand. Daily demand follows a Poisson process with a weekly seasonal pattern. The warehouse runs a base-stock (order-up-to) policy: at the end of each day it calculates the inventory position (on-hand plus pipeline orders minus any backlog) and orders enough to raise that position to a preset target. Orders arrive after the product’s lead time. When serving customers each day, the warehouse first clears any backlog and then serves new demand. Costs are recorded each day as holding plus backorder penalties, but only after an initial warm-up period used to stabilize the system.

\textbf{Objective.} The optimization task is to choose an integer order-up-to target for each product, within practical lower and upper bounds, to minimize the expected average operating cost over the planning horizon. Capacity can be handled either as a hard limit on the sum of all targets or as a soft constraint by adding an increasing penalty (for example, quadratic) when the total target exceeds a specified capacity. Under the chosen targets, the base-stock policy and fixed lead times fully determine material flows and, in turn, the expected costs under the seasonal, stochastic demand.

\textbf{Baseline SO algorithms.} We benchmark three BO variants and two population-based heuristics on the common bounded design space for the base-stock vector $\mathbf{S}$. BO employs a Gaussian process surrogate using a Matérn kernel. We employ (i) expected improvement (EI), (ii) upper confidence bound (UCB), and (iii) probability of improvement (PI) as the acquisition functions. Regarding the heuristics algorithm, the Genetic Algorithm (GA) uses a population of 10 with tournament selection, uniform crossover (rate 0.7) and per-coordinate Gaussian mutation (rate 0.2; step size $0.1$ of the variable range); all offspring are clipped to the bounds. The Particle Swarm Optimizer (PSO) maintains 5 particles, asynchronous one-particle updates, and bound clipping. All baselines consume the same evaluation budget and serve both as standalone solvers and as components in hybrid schedules.

\textbf{Experimental results.} All methods were evaluated under budgets of $100$ evaluations across $5$ independent runs. We exhibit the results including all 5 baseline SO algorithms and the top 3 proposed schedule of algorithms (see Table \ref{tab:mean-perf}). Among single baseline SO algorithms, \textbf{BO-PI} achieved the lowest mean cost ($28.20\!\pm\!1.22$), narrowly ahead of \textbf{BO-EI} and \textbf{BO-UCB}; \textbf{GA} was moderately worse on average, and \textbf{PSO} exhibited the largest variance and the worst mean under this budget. 

\begin{figure}
    \centering
    \includegraphics[width=0.8\linewidth]{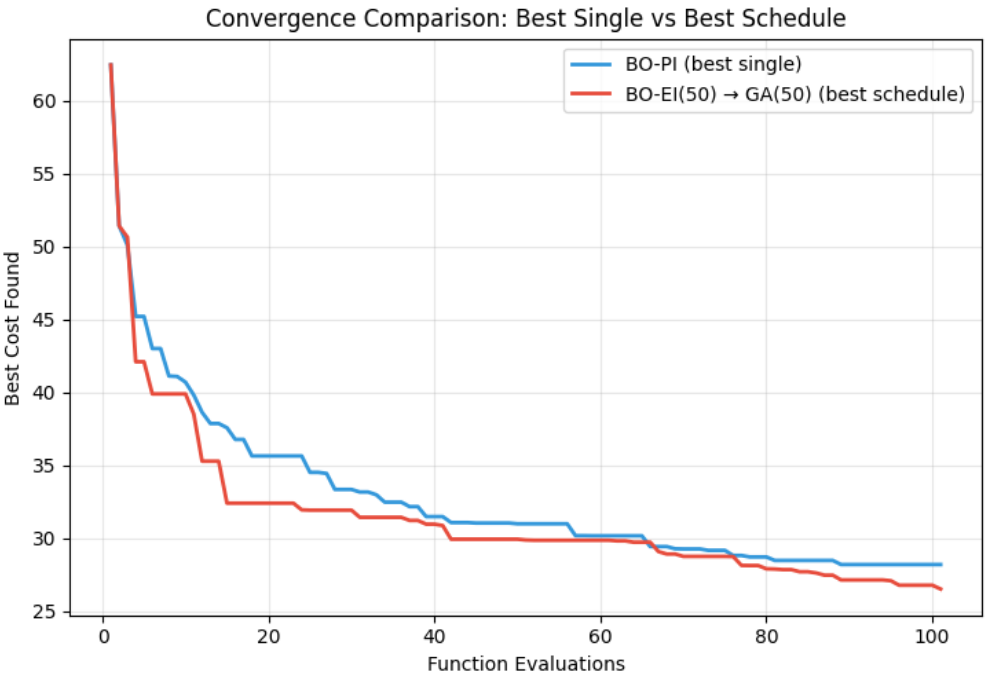}
    \caption{The best-so-far (minimal) values of the best baseline (BO-PI) and the best schedule (BO-EI(50) $\rightarrow$ GA(50)).}
    \label{fig:traj1}
\end{figure}

The schedule of hybrid SO algorithms improved both performance (lower optimal costs) and robustness (lower variance of the optimal costs): the schedule \textbf{BO-EI(50)$\rightarrow$GA(50)} delivered the best overall performance with mean cost $26.52\!\pm\!0.85$, a \textbf{6.0\%} reduction relative to the best single method. Qualitatively, the schedules showed faster early decrease in cost and tighter dispersion by exploiting the information from the previous iterations: evaluations accumulated by the first stage seed the second (e.g., GA’s population inherits high-quality designs), combining BO’s model-based search with GA’s refinement. Overall, BO variants are strong standalone algorithms for this problem class, while hybrid schedules yield lower final cost and reduced variability under fixed evaluation budgets. We also include the optimization trajectories (best-so-far plots) of the experiments in Figure \ref{fig:traj1}.

\begin{table}[t]
\centering
\caption{Average cost (lower is better) with one–standard-deviation error across $5$ runs; each method uses a budget of $100$ evaluations.}
\label{tab:mean-perf}
\begin{tabular}{lcc}
\toprule
Method & Mean $\pm$ Std. & Notes \\
\midrule
\multicolumn{3}{c}{\emph{Single algorithms}}\\
\midrule
BO-EI  & $28.27 \pm 1.16$ &  \\
BO-UCB & $28.50 \pm 0.89$ &  \\
\emph{BO-PI} & \emph{$28.20 \pm 1.22$} & Best single \\
GA     & $28.68 \pm 1.79$ &  \\
PSO    & $43.97 \pm 19.90$ & Highest variance \\
\midrule
\multicolumn{3}{c}{\emph{Hybrid schedules}}\\
\midrule
\textbf{BO-EI(50) $\rightarrow$ GA(50)} & \textbf{$26.52 \pm 0.85$} & Best overall \\
BO-UCB(25) $\rightarrow$ PSO(25) $\rightarrow$ BO-PI(50) & $27.44 \pm 1.34$ &  \\
GA(20) $\rightarrow$ BO-EI(40) $\rightarrow$ BO-PI(40)   & $27.58 \pm 1.21$ &  \\
\bottomrule
\end{tabular}

\vspace{0.25em}
\end{table}

\subsection{Multi-Server Queuing Network and Resource–Routing Optimization}\label{subsec:queuing-rr}
We consider a system in service operations. More specifically, the system is a multi-server queuing network with several service stations operating in discrete time. Customers arrive according to a Poisson process and are initially admitted to an entry station; upon service completion, a customer either exits the system or, with a prescribed probability, is routed to another station according to a row-stochastic routing matrix. Each station is staffed by an integer number of parallel servers drawn from a fixed system-wide pool, and its effective service rate can be boosted via a continuous “service-rate multiplier.” Unserved customers wait in first-come queues. At each period, we accrue (after a warm-up phase) holding cost proportional to total queue length, operating cost per active server, and resource cost for service-rate enhancements; additional penalties discourage excessive congestion and load imbalance across stations.

\textbf{Objective.} The optimization task is to choose (i) an integer server allocation at each station subject to a total-server budget, (ii) continuous service-rate multipliers within prescribed bounds, and (iii) routing probabilities whose rows normalize to one, so as to minimize the expected average operating cost over the horizon. Under any candidate configuration, the induced stochastic dynamics (arrivals, services, and routing) fully determine customer flows and thus the resulting holding, server, resource, and penalty costs.

\begin{figure}
    \centering
    \includegraphics[width=\linewidth]{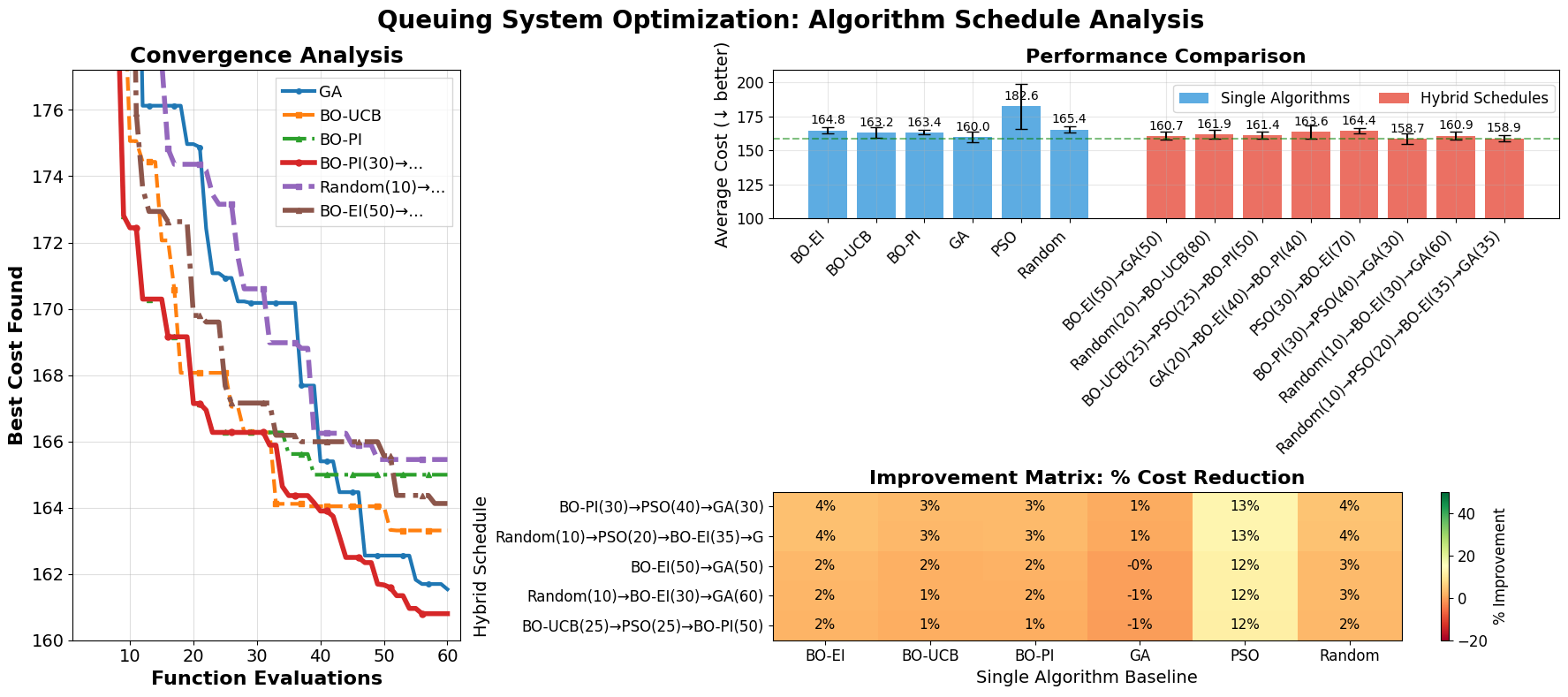}
    \caption{The experiments for the multi-server queue.}
    \label{fig:exp2}
\end{figure}

The experimental results are summarized in Figure \ref{fig:exp2}, which deliver similar results as in Sec.~\ref{subsec:warehouse-bs}, and therefore support the superiority of SO schedules (as well as SOCRATES) over standalone algorithms.

\section{Related Work \& Comparison}
We discuss some related work of our proposed SOCRATES, and compare our method with existing literature.

\subsection{Digital Twins}
Our method is connected to and benefits from digital twins (DTs).
DT has emerged as a powerful paradigm, creating high-fidelity virtual replicas of physical assets, processes, and systems to enable advanced simulation, prediction, and optimization. By establishing a dynamic, data-driven virtual replica of a real-world system, a DT serves as a sophisticated platform for applying operations research (OR) methodologies to understand past events, predict future outcomes, and prescribe actions to solve complex problems before they occur \cite{biller2023practitioner}. This capability is particularly transformative for managing large-scale stochastic systems, with a prominent application being the Supply Chain Digital Twin (SCDT) \cite{srai2019supply,prockl2025supply}. An SCDT models the supply network as a spatio-temporal dynamic system, using real-time data to enable unprecedented visibility and agility. This allows for the application of OR techniques to pressing supply chain challenges, such as using high-fidelity simulations for resilience testing and risk management, and employing real-time optimization to dynamically control inventory, logistics, and production in response to disruptions.

To better distinguish our proposed digital replicas from the current implementation of DTs, we refer to them as Operational AI Replicas (OARs), emphasizing that they are purpose-built for operational evaluations (i.e., evaluating and revising optimization algorithms) and are constructed using AI techniques. There are mainly three differences between conventional DTs and our OARs:

\begin{itemize}
\item \textbf{Purpose-Built for Algorithm Development:} OARs are specifically designed as an ``algorithm workbench'' for the offline development, evaluation, and revision of optimization algorithms. In contrast, the primary focus of conventional DTs is often on creating a high-fidelity, real-time replica of a physical system for live monitoring and control.

\item \textbf{Automated and Cost-Effective Construction:} Our OARs are built through a highly automated process using LLMs and an EM-type learning procedure, which is significantly less expensive than traditional methods. This differs from many DTs that require extensive, resource-intensive efforts from human experts to manually model physical systems.

\item \textbf{Scalability and Computational Efficiency:} By design, OARs are computationally efficient and scalable, which is essential for their task of running numerous simulations to evaluate algorithms. This is a key advantage, as many high-fidelity DTs can be resource-intensive, demanding substantial computational power to maintain real-time synchronization with their physical counterparts.
\end{itemize}

Compared to DT, our proposed OARs are more analogous to digital cousins (DCs) proposed in \cite{dai2024automated}. The defining characteristic of a DC is the deliberate relaxation of the ``exact replica'' requirement that defines a DT. A DC does not aim to model every detail of its real-world counterpart. Instead, it focuses on preserving higher-level, functionally important properties. The goal is not perfect imitation but the creation of a distribution of similar environments to train more robust and generalizable policies. While OARs are conceptually similar to DCs, the primary difference lies in the domain of abstraction: DCs are geometric and visual abstractions. They are built from an RGB image and a 3D asset library, resulting in a visually and physically interactive simulation. OARs are causal and analytical abstractions. They are built from textual descriptions and I/O data, resulting in computational efficient functions/mappings that compose stochastic systems.

\subsection{LLMs for Optimization Modeling}
\label{sec:llmforopt}
The intersection LLMs and mathematical optimization is a rapidly advancing field aimed at addressing a challenge in real-life application of optimization: the specialized expertise required to translate real-world problems into formal mathematical models. This emerging synergy leverages LLMs not as direct solvers, but as ``optimization modeling copilots'' that can interpret natural language descriptions, formulate mathematical models, and generate executable code for traditional high-performance solvers. The primary goal is to democratize access to powerful optimization tools, enabling domain experts without formal optimization modeling training to tackle complex decision-making challenges \cite{abdel2025teaching}. 

A core challenge is the translation of ambiguous natural language into the precise structure of a mathematical programming \cite{roth2017integer}. Early efforts, such as the NL4Opt competition \cite{nl4opt2022}, formalized this task by breaking it down into identifying key components (decision variables, objectives, and constraints), and generating a logical representation. To improve reliability, leading frameworks have adopted structured intermediate representations as a crucial bridge between language and solver code. These representations, such as a formal mathematical model in LaTeX, force the LLM to create a clear and verifiable specification before generating code. The accuracy of this formulation is further enhanced by sophisticated prompt engineering techniques, most notably Chain-of-Thought (CoT) prompting \cite{wei2022chain}, which guides the LLM through a step-by-step reasoning process. 

As the field has matured, several specialized architectural frameworks have been developed, each with a distinct philosophy. ORLM \cite{orlm2024} represents a data-centric approach, utilizing a semi-automated data synthesis engine called OR-Instruct to create vast, diverse datasets for fine-tuning smaller, open-source models. This has enabled models with 7-8 billion parameters to achieve state-of-the-art performance, demonstrating the power of domain-specific training while addressing cost and privacy concerns. In comparison, OptiMUS \cite{ahmaditeshnizi2024optimus} employs a modular, multi-agent architecture to handle large-scale, complex problems. It decomposes the modeling process into tasks for specialized agents (Formulator, Programmer, Evaluator) and uses a ``connection graph'' to manage context, thereby overcoming the input length limitations of LLMs. Additionally, LLMOPT \cite{jiang2025llmopt} is a unified framework focused on improving ``optimization generalization.'' It integrates its universal five-element formulation with multi-stage learning, including model alignment to reduce hallucinations and an automated self-correction loop to refine solutions. 

The final stage of the pipeline involves code generation and execution, where the ``solver-in-the-loop'' paradigm has become critical. These proposed LLM frameworks not only generate Python code for solvers like Gurobi but also execute it, interpret feedback (such as runtime errors), and autonomously debug their own code or even the underlying mathematical model. The efficacy of these systems is measured using specialized benchmarks like NL4OPT \cite{nl4opt2022}, MAMO \cite{huang2024llms}, and the industrially-focused IndustryOR \cite{orlm2024}. Performance metrics such as solving accuracy and execution rate consistently show that these specialized, fine-tuned frameworks outperform general-purpose models like GPT-4.
\end{document}